# Leveraging Human Reading Behavior for Keyphrase Extraction: A Webcam-based Eye-tracking Corpus

Chengzhi Zhang*, Xinyi Yan, Wenqi Yu

Department of Information Management, Nanjing University of Science and Technology, Nanjing, 210094, China

{zhangcz, yanxinyi, yuwq}@njust.edu.cn

**Abstract**: **Purpose –** Keyphrases are not only statistically or semantically important textual units, but also words or phrases that are more likely to attract readers' attention during comprehension. However, existing KPE studies have mainly focused on improving models' representation learning from input texts, while largely overlooking the connection between keyphrases and human reading behavior. This study investigates whether lightweight webcam-based eye-tracking features can enhance KPE from Chinese academic abstracts in Library and Information Science (LIS).

**Methodology/Approach -** Motivated by the limited availability of eye-tracking data for Chinese academic reading, we developed a lightweight webcam-based data collection platform by integrating the open-source SearchGazer library. Based on the collected and preprocessed data, we constructed the Chinese LIS Eye-Tracking Corpus (CLIS-ET). We then incorporated three character-level eye-tracking features, first fixation duration (FFD), fixation number (FN), and total fixation duration (TFD), into KPE models to evaluate their effects on extraction performance.

**Findings -** By integrating eye-tracking features into KPE models, our experiments demonstrated that the combination of fixation number and total fixation duration (FN+TFD) achieved the best performance on the Att-BiLSTM+CRF-based KPE model, highlighting the significant impact of eye-tracking data on optimizing keyphrase extraction from academic literatures.

**Originality/value -** This paper presents a cost-effective eye-tracking methodology to improve keyphrase extraction from academic papers. We also introduce the Chinese Academic Eye-Tracking Corpus, which includes key eye-tracking metrics: fixation number, first fixation duration, and total fixation duration. Our findings demonstrate that incorporating these features enhances KPE models, with first fixation duration yielding the most significant performance improvement. The dataset and source code can be accessed at:

* Corresponding author: Chengzhi Zhang (zhangcz@njust.edu.cn).

https://github.com/yan-xinyi/ET_AKE.

# 1. Introduction

Keyphrase extraction (KPE) is a critical task in information retrieval, helping users efficiently navigate academic literature. Existing KPE methods mainly identify keyphrases based on text-internal cues, such as term frequency, position, syntactic patterns, and contextual semantics, but they usually pay limited attention to how readers attend to and process important textual units during comprehension. However, eye-tracking studies have shown that fixation patterns are closely associated with attention allocation and cognitive processing during reading (Rayner, 1998). Recent studies further suggest that gaze-derived features can provide useful auxiliary signals for text understanding tasks, including sentiment analysis, relation classification, and KPE (Hollenstein et al., 2019; Zhang and Zhang, 2021; Yan et al., 2024). These studies integrate behavioral insights into human attention and cognition during reading, providing a deeper understanding of reading processes that can enhance KPE.

Eye-tracking, established over the past thirty years as a dependable measure of attentional focus and cognitive processing, provides an in-depth view of cognitive behavior when engaging with text (Ma et al., 2022). By analyzing gaze patterns, researchers can identify the sections of text that capture readers' attention most, reflecting psychological processes linked to comprehension and learning. Eye-tracking has applications across psychology, linguistics, and human-computer interaction, where it reveals how individuals engage with computers in cognitive tasks (Harm & Seidenberg, 2004; Chen et al., 2023). However, traditional eye-tracking studies often require strict experimental controls, which limits their broader applicability. Recent advancements in eye-tracking data acquisition and processing technology have led to the construction of open-source eye-tracking corpora, including ZuCo (Hollenstein et al., 2018), GECO (Cop et al., 2017), and Dundee (Kennedy, 2003). These resources have reduced the costs of obtaining eye-tracking data, expanding its potential applications.

With these technological advancements, eye-tracking has become more accessible, enabling broader exploration in real-world contexts. Open-source corpora have facilitated studies that integrate eye-tracking data into natural language processing (NLP) tasks, including sentiment analysis, part-of-speech tagging, and KPE (Barrett et al., 2016; Hollenstein et al., 2019). During reading, individuals focus on key text segments, reflecting cognitive processing patterns that align with KPE objectives (Rayner, 1998). This connection suggests that human-computer interactions in reading may enhance computational models of KPE through the integration of psychological insights.

Despite these advancements, existing reading eye-tracking corpora still provide limited support for academic and domain-specific texts. This limitation is particularly relevant to KPE, because disciplinary terminology, abstract structure, and readers' prior knowledge may influence how attention is allocated to potential keyphrases during reading. In the Chinese context, such resources are even more limited. To address this gap, this study focuses on Library and Information Science (LIS), a field with domain-specific terminology and well-established academic abstract conventions, and constructs the Chinese LIS Eye-Tracking Corpus (CLIS-ET). This corpus captures three critical eye-tracking metrics—first fixation duration (FFD), fixation number (FN), and total fixation duration (TFD)—enabling the analysis of how reading-behavior signals relate to KPE performance. Building on Zhang & Zhang's (2021) approaches, utilizing eye-tracking data as ground truth for neural network attention mechanisms or as external features for KPE, our research incorporates eye-tracking features into multiple KPE architectures.

Ultimately, this study contributes to the understanding of eye-tracking's role in KPE, framing it as a window into cognitive interactions in academic reading. This perspective emphasizes the psychological dimensions of human-computer interaction and its potential to enhance information retrieval tools.

In summary, this study presents the following three contributions:

First, we develop a lightweight webcam-based data collection platform for collecting reading-behavior signals during the reading of Chinese LIS academic abstracts. The source code for the reading eye-movement data collection platform is available at: https://github.com/yan-xinyi/Reading_ET_System. The platform integrates open-source web technologies and consumer-grade webcams, reducing hardware and deployment requirements for domain-specific eye-tracking data collection.

Second, we introduce the Chinese LIS Eye-Tracking Corpus (CLIS-ET), a domain-specific

corpus that includes three core eye-tracking features: fixation number (FN), first fixation duration (FFD), and total fixation duration (TFD). The corpus provides reading-behavior evidence for studying cognitively grounded KPE in this disciplinary context.

Third, we evaluate the effects of individual and combined eye-tracking features across different KPE architectures. The results show that eye-tracking features provide complementary cognitive signals for KPE, and the significance analysis across multiple models further demonstrates their broadly consistent contribution across multiple KPE models.

# 2. Related Work

The research relevant to this paper mainly focuses on the construction of eye-tracking corpora, the application of eye-tracking data in NLP, and keyphrase extraction methods.

## 2.1 Construction of Eye-tracking Corpora

**Table 1. Overview of established eye-tracking corpora**

| Authors | Corpus name | Language | Corpus source | Corpus scale (#Sentences) | Sampling Equipment and Frequency |
|---|---|---|---|---|---|
| *Kennedy et al., 2013* | Dundee | English, French | Independent Newspaper, World Newspaper | 2379 | Dr.Bouis Oculometer eye-tracking device (1000Hz) |
| *Kliegl et al., 2004* | Potsdam | German | Daily German Short Sentences | 144 | SR EyeLink System eye-tracking device (250Hz) |
| *Luke & Christianson, 2018* | Provo | English | News, Magazines, and Novels | 134 | SR Research EyeLink 1 000 Pluseye-tracking device (1000Hz) |
| *Cop et al., 2017* | GECO | English, Dutch | Novel "The Mystery of Thiels Manor" | 5242 | EyeLink 1000eye-tracking device (1000Hz) |
| *Hollenstein et al., 2018* | ZuCo | English | Stanford Sentiment Treebank, Wikipedia Relation Extraction Corpus | 1107 | EyeLink 1000 Plus eye-tracking device(500Hz)[1] |
| *Sui et al., 2023* | GECO-CN | Chinese, English | Novel "The Mystery of Thiels Manor" | 5242(English),5 066(Chinese) | EyeLink 1000 Plus eye-tracking device (1000Hz) |
| *Pan et al., 2022* | Beijing Sentence Corpus (BSC) | Chinese | Simple Chinese Sentences | 120 | EyeLink II (500Hz) |
| *Siegelman et al., 2022* | MECO | English German and 13 other languages | Wikipedia | 110 for each language | EyeLink Portable Duo, SR Research and other eye-tracking devices[2] |
| *Zhang et al., 2022b* | Chinese Eye Movement Measurement Database | Chinese | Natural Chinese Sentences | 8015 | EyeLink 1000 eye-tracking device |

Since the 1980s, the significance of eye-tracking data has been acknowledged, attracting considerable attention to the construction of eye-tracking datasets (Rayner, 1998). Reading

1 https://www.sr-research.com/eyelink-1000-plus-technical-specifications
2 https://www.sr-research.com/eyelink-portable-duo

eye-tracking corpora, meticulously crafted by experts, represent a form of multimodal corpus. Prominent international reading eye-tracking corpora typically encompass details such as corpus content, word length, and fixation metrics (Wang & Zhao, 2020). Certain studies also evaluate participants' language cognitive abilities through various tests (Sui et al., 2023; van der Sluis & van den Broek, 2023) . The acquisition of eye-tracking data adheres to a standardized procedure, comprising stages such as calibration, formal reading, in-reading, and post-reading questioning. Eye-tracking metrics commonly refer to attributes arising from fixations, saccades, and regressions. The most prevalent features include single fixation duration, first fixation duration, and total fixation duration.

Table 1 presents a list of commonly used eye-tracking corpora for reading. A comparison of the key characteristics of these datasets reveals that, although the number and scale of eye-tracking corpora have grown in recent years, the experimental texts are primarily sourced from microblogs, news articles, and novels, with a noticeable lack of academic literature. As a result, insights from these corpora may not generalize well to academic texts, limiting progress in related research. Additionally, the high cost of specialized eye-tracking equipment further restricts large-scale data collection. To address these challenges, our study introduces a lightweight and accessible way to collect eye-tracking signals using widely available webcams, supporting scalable data collection for Chinese LIS academic reading and supporting future applications of eye-tracking research in academic reading.

We further summarize the hardware requirements, deployment conditions, sampling rates, experimental environments, accessibility, and approximate costs of commonly used eye-tracking tools in Appendix A.1. As shown in the comparison, laboratory-grade systems provide high sampling rates and are widely used in controlled eye-tracking experiments, but they usually require specialized hardware, calibration procedures, and laboratory-based deployment. In contrast, webcam-based eye-tracking solutions rely on consumer-grade cameras and standard computing devices, making them more lightweight and accessible for large-scale or remote data collection. Recent validation studies also support the practical value of webcam-based eye tracking. Kaduk et al. (2024) compared a webcam-based system with the EyeLink 1000 and reported that although webcam-based systems should not be regarded as direct substitutes for professional eye trackers in high-precision oculomotor experiments, they provide a feasible and accessible option for collecting general reading-behavior signals.

## 2.2 Application of Eye-Tracking Data in Natural Language Processing

In recent years, eye-tracking technology has been widely used in NLP due to its ability to reveal users' reading patterns and attention distribution, providing valuable cognitive insights for text-processing tasks. Bačić & Henry (2022) used eye-tracking to assess cognitive effort in cognitive fit theory, highlighting fixation metrics and physiological indicators for deeper insights, offering valuable implications for NLP research. Van der Sluis & van den Broek (2023) found that text relevance influences reading behavior, as reflected in differences in eye-tracking measurements. As a result, eye-tracking data is often leveraged to enhance various NLP applications.

In the field of text summarization, Xu et al. (2009) introduced a personalized document summarization algorithm that relies on individual users' attention time, captured through vision-based eye-tracking, to predict the level of attention given to each word in a document. Klerke et al. (2016) employed eye-tracking data from the Dundee corpus to optimize sentence compression tasks on news texts. Barrett et al. (2016) employed the same eye-tracking corpus to improve weakly supervised part-of-speech tagging tasks. In recent years, research on machine translation leveraging eye-tracking data has garnered increasing attention. Du et al. (2024) explored the integration of eye-tracking technology into real-time translation systems to enhance reading experiences, demonstrating that eye-tracking-assisted translation reduces cognitive load and improves comprehension.

In sentiment analysis and information extraction, eye-tracking features have proven useful as well (Zhao et al., 2023). Mishra et al. (2016) incorporated eye-tracking features into traditional models for sentiment analysis of user-generated content. Hollenstein et al. (2019) demonstrated that first fixation duration improves the performance of relation classification, sentiment recognition tasks, and named entity recognition. Xie et al. (2021) leveraged eye-tracking to provide cognitive insights relevant to attention modeling in semantic analysis tasks. Additionally, Jin et al. (2023) employed eye-tracking experiments to reveal differences in consumers' attention allocation when reading different types of reviews, thereby advancing the understanding of semantic processing mechanisms in micro-blogs and providing an empirical foundation for NLP models to more accurately simulate human semantic perception.

Similarly, some researchers have utilized user search behavior for KPE, as demonstrated by AKEGIS (Scholz et al., 2019), which extracts keywords from internal search logs, providing a downstream application reference for leveraging eye-tracking data in keyword extraction research. Zhang & Zhang (2021) verified that total fixation duration from the ZuCo Corpus enhances the performance of KPE on microblogs. Yan et al. (2024) further explored the impact of eye-tracking and EEG features from the ZuCo Corpus on the performance of KPE from Weibo, achieving additional improvements. However, the ZuCo Corpus used in their study is based on general Wikipedia content and does not provide eye-tracking information for Chinese academic vocabulary. To address this gap, our study proposes an efficient eye-tracking data collection approach capable of simultaneously collecting data from multiple participants. We also construct an eye-tracking dataset and investigate the impact of various eye-tracking features and their combinations on KPE.

Recent studies indicate that eye-tracking can bridge cognitive attention patterns and computational keyword identification, providing a behavioral foundation for salience modeling in KPE. Eye-tracking features—particularly fixation duration and frequency—reflect the depth of semantic processing and the importance of lexical items, which aligns closely with the objectives of KPE. Incorporating such cognitive indicators into models helps move beyond surface-level statistical methods by simulating how readers naturally focus on key information during comprehension, thereby improving the identification of core concepts. The integration of cognitive psychology and natural language processing highlights the potential of eye-tracking as a powerful complementary signal to enhance the interpretability and performance of KPE models.

### 2.3 Methods of Keyphrase Extraction

The effectiveness of keyphrase extraction from academic texts directly impacts downstream tasks such as academic topic analysis and text recommendation, making it a topic of significant research value. Based on the review studies by Nasar et al. (2019) and Xie et al. (2023), KPE methods are generally categorized into unsupervised and supervised approaches. The representative studies are summarized in Table 2.

Unsupervised methods primarily identify candidate keyphrases based on statistical features and word graph structures. These statistical features include word weight attributes such as frequency, part of speech, and length, as well as positional information. YAKE (Campos et al., 2020), for example, represents a purely statistical approach that extracts keywords using multiple single-document features such as casing, word position, and term frequency, without relying on external corpora or linguistic annotations. Graph-based approaches excel in processing short texts or low-density documents (Vega-Oliveros et al., 2019) but often struggle with complex semantic relationships due to their reliance on structural patterns within the text. To address these limitations, recent studies have incorporated richer linguistic and semantic features into traditional graph-based frameworks. For instance, HSN (Hierarchical Semantic Network) (Yoo et al., 2020) leverages hierarchical relationships and centrality measures to address the limitations of traditional graph-based methods in capturing complex semantic connections. Duari et al. (2019) showed that competitive performance across low-resource languages without relying on external tools, while AdaptiveUKE (Liu et al., 2024) improves topic coherence by introducing a gated topic modeling mechanism for more adaptive keyphrase weighting. Xu et al. (2025) proposes a shallow-to-deep ranking framework to refine candidate selection across hierarchical feature representations. Additionally, MGRank (Goz et al., 2022) reinforces graph-based methods by integrating deeper semantic structures, improving their capacity to capture intricate dependencies within texts. These advancements leverage transformer-based architectures to enhance keyphrase extraction, bridging the gap between statistical heuristics and deep learning-driven strategies.

**Table 2. Summary of representative studies of KPE**

| Categories | Authors | Methods | Main findings |
|---|---|---|---|
| Unsupervised methods | *Vega-Oliveros et al., 2019* | *Multi-Centrality Index(MCI)* | MCI outperforms standalone centrality and other clustering algorithms. |
| | *Yeon & Kim, 2020* | *Hierarchical Semantic Network (HSN)* | The proposed method outperformed existing network-based KPE methods in selecting representative topical keyphrases. |
| | *Duari et al., 2019* | *sCAKE* | sCAKE can capture semantic connectivity between words, which significantly improves keyword quality over existing graph-based methods |
| | *Campos et al., 2020* | *YAKE* | YAKE outperforms existing methods across multilingual and cross-domain tasks without relying on external resources. |
| | *Liu et al., 2024* | *AdaptiveUKE* | AdaptiveUKE surpasses state-of-the-art models across several benchmarks, achieving top performance on six keyphrase extraction datasets. |
| | *Xu et al., 2025* | *SDRank* | SDRank consistently outperforms other robust unsupervised models, emphasizing the advantages of integrating diverse semantic relationships in unsupervised keyphrase extraction. |
| Supervised methods | *Conde et al., 2016* | *LiTeWi* | LiTeWi outperforms Point-by-Point Mutual Information (PMI), Cardinality (X2), and Wikifier in performance. |
| | *Xie et al., 2017* | *Sequential pattern mining algorithm* | The new approach outperforms existing methods in terms of both runtime efficiency and completeness. |
| | *Duari & Bhatnagar, 2020* | *Complex Network* | The proposed framework outperforms recent techniques with significant results on scientific and news corpora. |
| | *Yang et al., 2022* | *GCN-based representation* | The GCN-based method surpasses previous state-of-the-art baselines on Micro-blog, Twitter, and StackExchange datasets. |
| | *Song et al., 2024* | *HybridMatch* | HybridMatch combines representation-focused and interaction-based matching modules for open-domain web keyphrase extraction. |

Supervised KPE methods can be broadly divided into machine learning and deep learning approaches. Common machine learning methods include Naive Bayes (Witten et al., 1999), Support Vector Machines (Mizuka et al., 2017), and Conditional Random Fields (Zhang et al., 2008). Recently, with advancements in deep learning, models like RNN, BiLSTM, and BiLSTM+CRF have been increasingly applied to keyphrase extraction from academic texts (Garg et al., 2020; Zhang & Zhang, 2019). Pre-trained language models, particularly BERT and RoBERTa (Gero & Ho, 2021), have been central to recent keyphrase extraction research. Approaches such as BERT-LSTM (Muralitharan et al., 2024), JointKPE (Martinc et al., 2022) and the GCN-based method (Yang et al., 2022) have significantly advanced the prediction of non-entity keyphrases by effectively utilizing contextual representations of the text. In general, KPE models based on deep learning exhibit significantly superior performance compared to machine learning models. Therefore, this study utilizes two types of deep learning models to extract keyphrases from academic texts and incorporates eye-tracking features from the developed eye-tracking dataset into the feature vectors, examining the influence of eye-tracking features on the performance of keyphrase extraction from academic texts.

Overall, although existing studies have demonstrated the potential value of eye-tracking features for KPE, most rely on expensive equipment and English-language corpora, making it difficult to extend their findings to Chinese academic contexts. The lack of low-cost, multi-participant Chinese academic eye-tracking data has limited progress in this field. Therefore, it is essential to develop more eye-tracking corpora and modeling frameworks to advance KPE research based on human reading behavior.

# 3. Methodology

## 3.1 Framework

The framework of this study is illustrated in Figure 1. First, we developed a Flask-based[3] data collection platform by integrating SearchGazer[4], an open-source JavaScript eye-tracking library that uses common webcams to infer users' gaze behavior in real time. Written in JavaScript, SearchGazer can be easily embedded into the browser with only a few lines of code. After data preprocessing and feature extraction, we constructed the Chinese Library and Information Science Eye-Tracking Corpus (CLIS-ET). Next, we develop a KPE model and integrated them with the eye-tracking features. Finally, we explore the impact of individual eye-tracking features and their combinations on the KPE task. The subsequent sections will follow the progression outlined in the research framework.

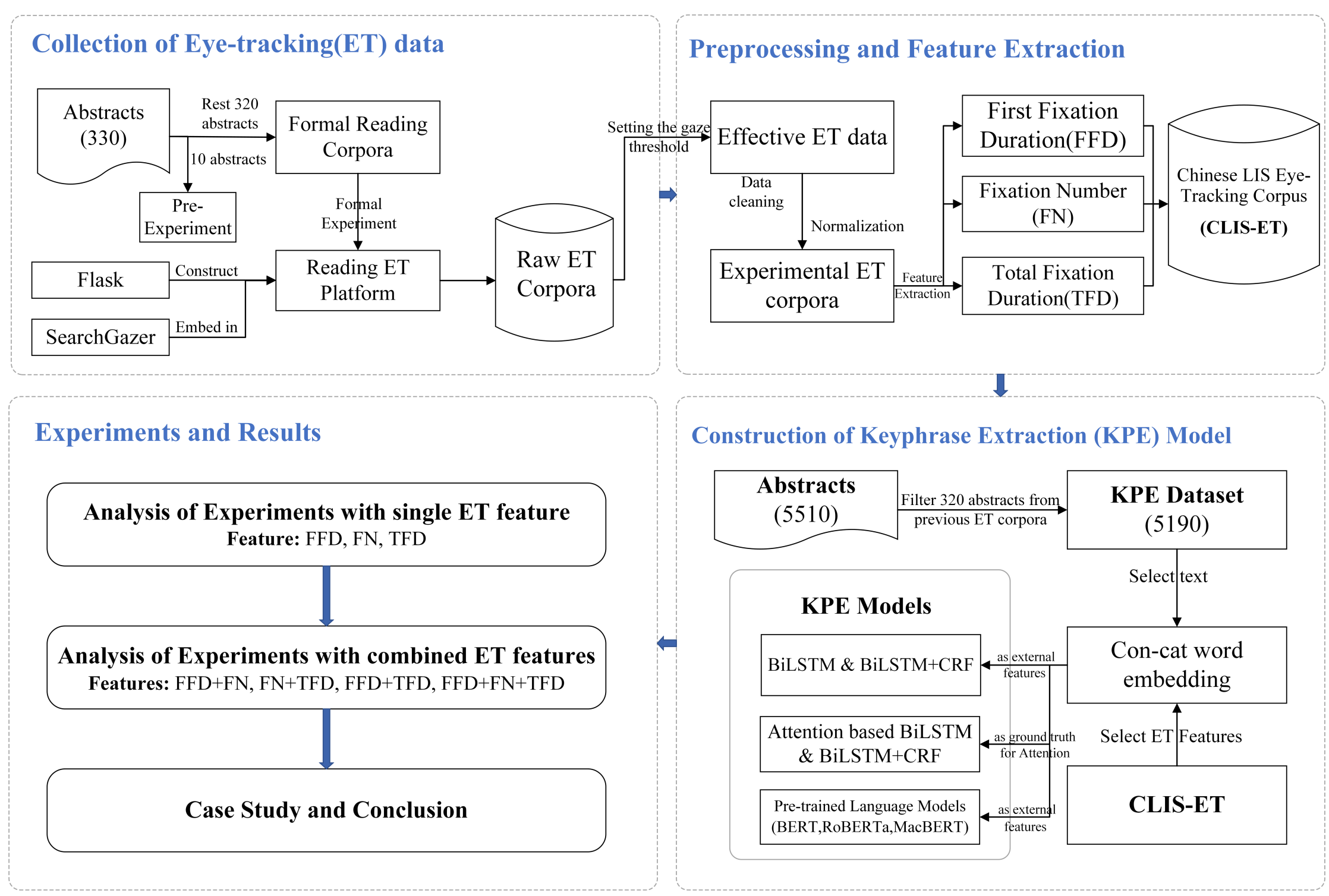

**Figure 1. Framework of this study**

3 Flask is a popular Python web framework used for building web applications. More information can be found at https://flask.palletsprojects.com/en/2.3.x/

4 SearchGazer：https://webgazer.cs.brown.edu/search/

### 3.2 Webcam-based Eye-Tracking Data Collection

#### 3.2.1 Reading Corpora for Eye-tracking Experiment

One of the main objectives of this paper is to create an accessible reading dataset for Chinese LIS literature. The data collection experiment involved ten graduate students or senior undergraduates from the field of Library and Information Science (LIS).

**Table 3. Description of the journals for the construction of eye-tracking reading corpora**

| Types of corpora | Source Journals | # Abstracts |
|---|---|---|
| Pre-Experiment | *Journal of the China Society for Scientific and Technical Information* (in Chinese) | 10 |
| Formal Experiment | *Journal of the China Society for Scientific and Technical Information* (in Chinese) | 100 |
| | *Information Science* (in Chinese) | 110 |
| | *Data Analysis and Knowledge Discovery* (in Chinese) | 110 |

A sample of Chinese academic articles published between 2000 and 2022 was selected, as outlined in Table 3. Notably, all the journals listed in the table are core journals in the field of LIS. Ten abstracts from *Journal of the China Society for Scientific and Technical Information* (in Chinese) were used for a pilot study, while the remaining abstracts were employed in the formal eye-tracking experiment. The final reading material includes 1,215 complete sentences (including titles), totaling 64,969 characters.

#### 3.2.2 Eye-Tracking Data Collection

The eye-tracking data collection experiment followed a carefully structured process to ensure reliability and consistency. The procedure included obtaining informed consent, explaining the experiment in detail, hands-on training, a preliminary trial, and the main data collection phase. Before the formal experiment, participants completed a pre-experiment session using ten randomly selected reading materials. This step helped them become comfortable with the setup while also ensuring stable data quality. The main experiment consisted of three key components: participant preparation, experimental environment setup, and eye-movement data collection. The following sections outline each component in detail.

(1) Participants

The number of participants was decided after reviewing existing eye-tracking datasets, which typically include approximately ten participants. Given that eye-tracking data is rich and multidimensional, a dataset of this size is sufficient to capture meaningful cognitive reading patterns while balancing feasibility and cost efficiency.

(2) Environment

The ten participants were divided into two groups, each completing the reading task under identical conditions and within the same timeframe. All experiments were conducted using the built-in webcams of laptop computers. To reduce potential variations caused by differences in hardware performance or individual physiological factors, participants first completed a 30-minute pre-experiment session. During this time, participants read the informed consent form, familiarized themselves with the procedure, and configured the experimental equipment.

(3) Eye movement collection

Typically, a single eye tracker can only gather data from one participant at a time, significantly limiting data collection efficiency. Thus, this study developed a user eye-tracking data collection platform based on the Flask framework[1] and the SearchGazer library[2], allowing real-time synchronization of users' eye movement coordinates during reading on the front end. The library supports a maximum sampling frequency of 60Hz for gaze coordinate collection, with an interval of approximately 16.67ms between consecutive sampling points. Previous research has defined a fixation as a visual pause lasting between 50ms and 1500ms, and the sampling frequency of the SearchGazer library is sufficient to identify these fixation segments. Further details regarding the experimental setup and participant recruitment can be found in A.2 Experimental Environment and Participants.

To maintain reading continuity, each reading task was structured as a complete abstract, with the first sentence displayed as the title. Before starting, participants were required to click on nine fixed points to calibrate their gaze position. Figure 2 illustrates the reading interface during the eye-tracking experiment. During the formal experiment, participants were instructed to keep their heads as still as possible, while the mouse cursor followed their gaze to further refine eye-tracking calibration. As reading time increased, fatigue could affect the accuracy of the eye-tracking data. To mitigate this, participants were allowed to take short breaks as needed. To ensure data reliability, a nine-point calibration was performed before each abstract, helping to maintain stable and precise gaze tracking. Once participants finished reading, they completed a brief test in which they selected the keyphrase of the abstract from a set of three options. This step was designed to gauge their level of focus during the task.

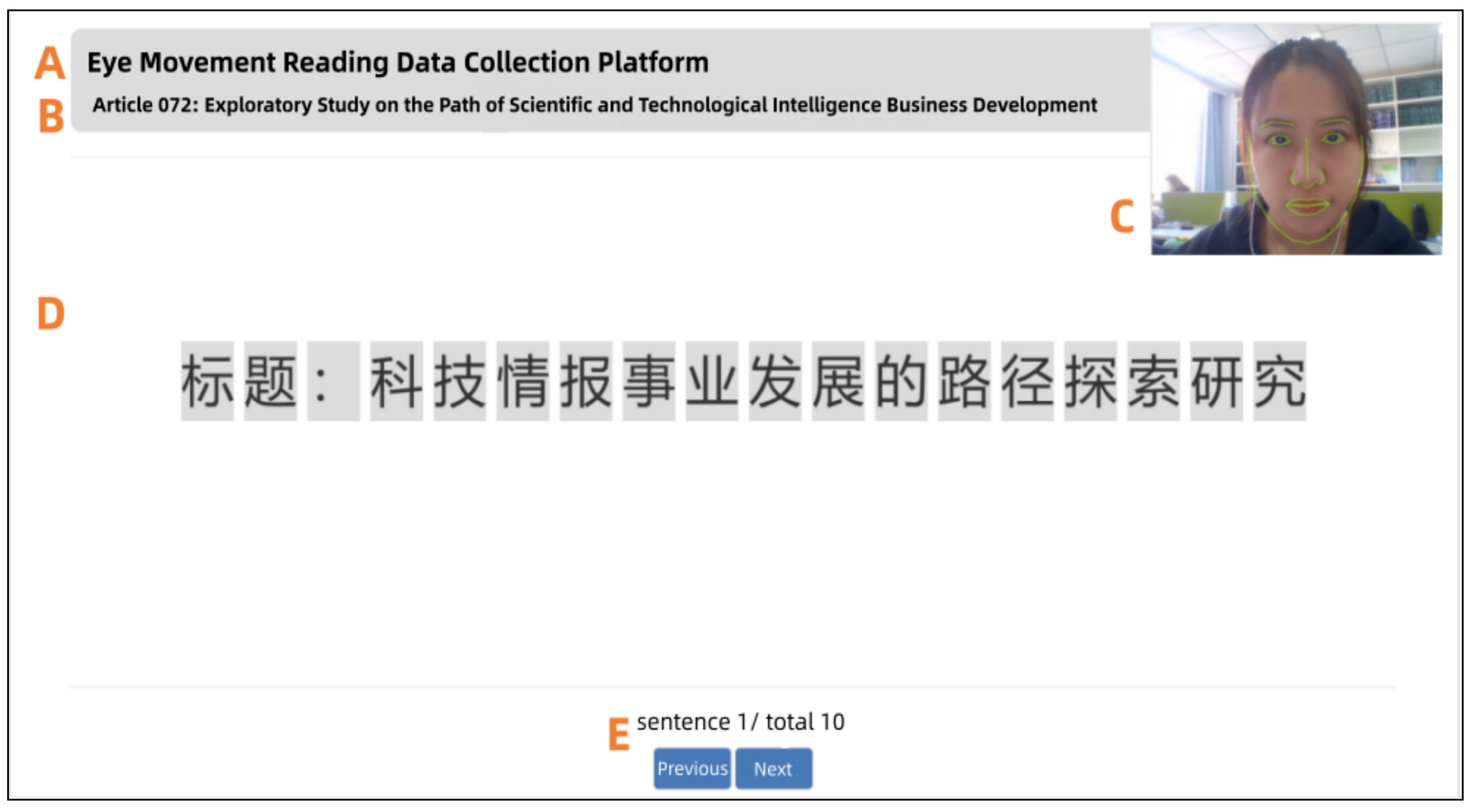

**Note:** ***A*** represents the name of the reading annotation platform, ***B*** indicates the article number and title corresponding to the content being read by the subject; ***C*** shows a facial alignment diagram of the reader, while ***D*** is the main reading interface, which means 'Title: Exploratory Study on the Development Path of Science and Technology Intelligence.'. The ***E*** indicates reading progress through an interactive design.

**Figure 2. Chinese academic text reading interface for eye movement experiments**

### 3.2.3 Preprocessing of Character-Level Eye-Tracking Data

This study records eye-tracking features at the character level, eliminating word segmentation issues and allowing for a more detailed analysis of reading processes. To preserve as much effective fixation data as possible, an effective fixation threshold of 16.7ms to 1500ms was set based on the sampling frequency of the SearchGazer eye-tracking library. This ensures that only fixation data within the range of 2 to 91 consecutive points on a single character are retained. The selected eye-tracking features include First fixation duration (FFD), Fixation number (FN), and Total fixation duration (TFD). Table 4 presents the definitions of these three eye-tracking measurement indicators.

To reduce the influence of individual differences in reading speed and habits, this study computes the average eye-tracking metrics for each character across participants. If more than half of the participants do not produce valid eye-tracking data for a given character, the eye-tracking feature value for that character is deemed invalid.

**Table 4. Definitions of eye-tracking measurement indicators**

| Eye-tracking Features | Abbreviations | Definitions |
|---|---|---|
| First fixation duration | FFD | Duration of first gaze on target character |
| Fixation number | FN | Total number of glances at the target character |
| Total fixation duration | TFD | Total gaze time on target character |

## 3.3 Chinese Library and Information Science Eye-Tracking Corpus (CLIS-ET)

Based on the collected and preprocessed webcam-based eye-tracking data, we constructed the Chinese Library and Information Science Eye-Tracking Corpus (CLIS-ET), a domain-specific corpus that records character-level reading behavior during the reading of Chinese LIS academic abstracts. The corpus contains three fixation-based features: first fixation duration (FFD), fixation number (FN), and total fixation duration (TFD). Despite the differences in reading speed among participants, the accuracy of the test questions was over 90%, suggesting that the participants remained attentive during reading.

To examine whether CLIS-ET captures cognitively meaningful signals for KPE, we compared the eye-tracking patterns of key and non-key characters. Table 5 summarizes the average eye-tracking features for all characters (All_Characters), key characters (Key_Characters), the top 50 most frequent key characters (Top50_Key_Characters), non-key characters (Non-key_Characters), and the top 50 most frequent non-key characters (Top50_Non-key_Characters).

**Table 5. Mean Character-Level Eye-Tracking Features in CLIS-ET (ms)**

| Character Category | FFD | FN | TFD |
|---|---|---|---|
| All_Characters | 33.6105 | 1.9809 | 69.1182 |
| Key_Characters | 34.0137 | 2.0614 | 73.4202 |
| Top50_Key_Characters | 34.1424 | 2.0484 | 72.9771 |
| Non-key_Characters | 33.3094 | 1.9436 | 66.8733 |
| Top50_Non-key_Characters | 32.0356 | 1.8694 | 63.9315 |

Table 5 shows that the average values for the three eye-tracking metrics are higher for key characters than for non-key characters, with the frequency of non-key characters having a greater impact on the eye-tracking feature values. This may be due to the high frequency of non-key characters which carry little actual meaning during reading. The differences in eye-tracking feature values between key characters and non-key characters are relatively small, possibly because the analysis only considered whether a character appeared in a keyphrase, without accounting for the probability of a character appearing in a keyphrase.

To address this limitation, the study selects characters from the entire set that belong to keyphrases, denoted as $P_{key}$. Equation (1) outlines the calculation of $P_{key}$. Where $N_W$ represents the frequency of a character's occurrence in the reading material, while $KN_W$ represents the frequency of a character's occurrence in the keyphrases of the material. As shown in Figure 3, we sorted 490 key characters by $P_{key}$ in ascending order and divided them into 9 groups based on the range of P values (since there are too few characters in the range of 0.7 to 1.0, only two groups were created). Subsequently, we calculated the mean eye-tracking features for key characters within each group.

$$P_{\text{key}} = \frac{KN_W}{N_W} \quad (1)$$

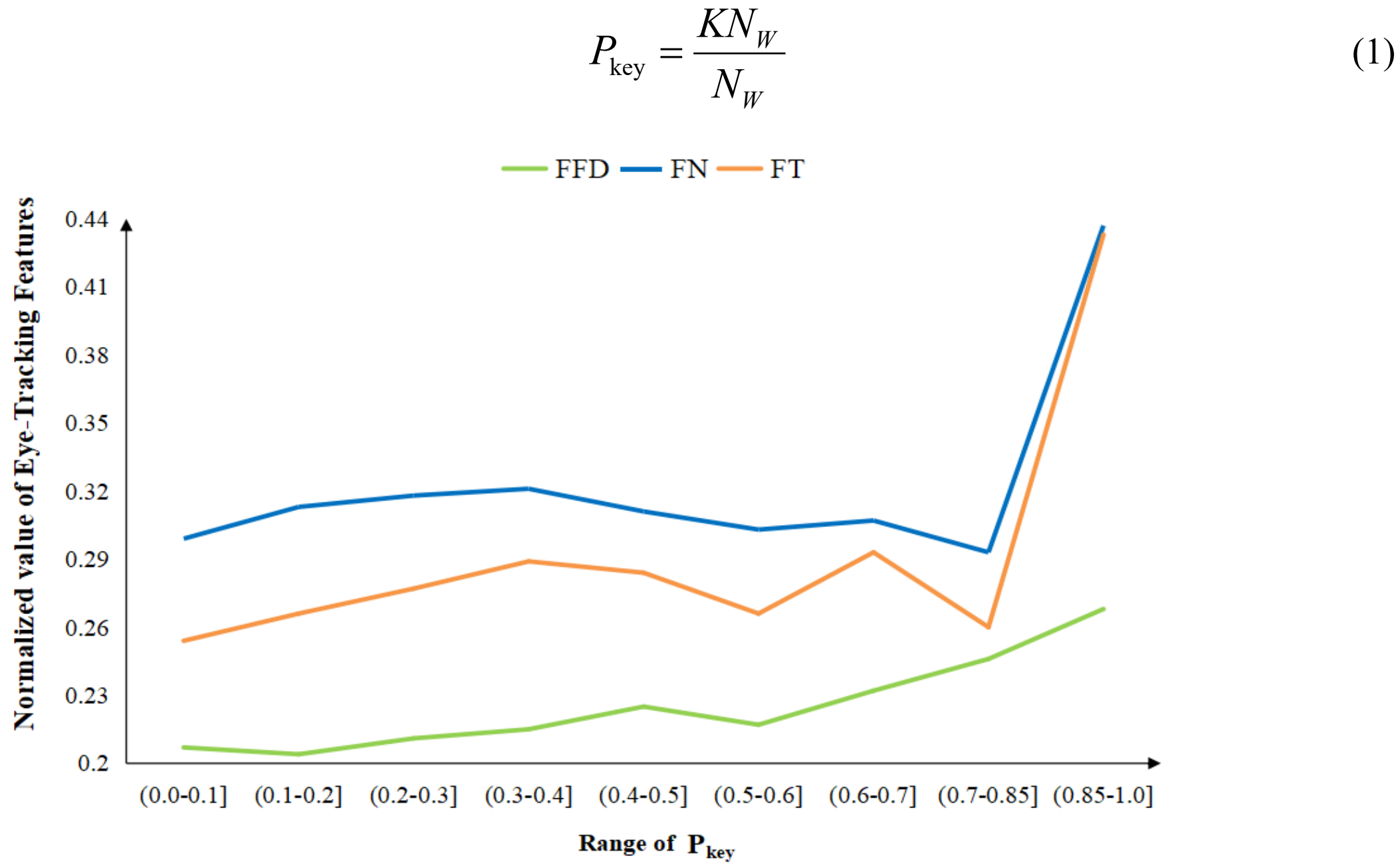

**Figure 3. Normalized Mean Eye-Tracking Features for Different $P_{key}$ Ranges**

It can be observed from Figure 3 that the FFD increases with the $P_{key}$ value. FFD is associated with the cognitive processing of initial reading; the more complex the cognitive process, the longer the fixation time. Thus, FFD also increases. FN and TFD follow similar patterns. When $P_{key}$ is less than 0.4, the mean values of the three eye-tracking features are positively correlated with $P_{key}$. When $P_{key}$ is between 0.4 and 0.85, the eye-tracking features fluctuate, peaking when the probability of a character appearing in a keyphrase nears 1. FN and TFD also reflect the number and duration of regressions to some extent. Research has shown that skilled readers exhibit higher saccade rates and fewer regressions compared to less skilled readers (Ashby et al., 2005). The variability in reading styles and levels among the participants in this experiment likely affects reading coherence and the number of regressions.

In summary, the constructed Chinese LIS Abstract eye-tracking dataset reveals differences in eye-tracking metrics between key characters and non-key characters. Additionally, the probability of a character appearing in a keyphrase correlates with higher eye-tracking feature values.

## 3.4 Construction of the Keyphrase Extraction Corpus

The reading materials for the eye-tracking experiment were selected from a compact corpus of 320 article titles and abstracts, collectively referred to as Abstract-320. Likewise, the keyphrase extraction dataset was built using titles and abstracts from the same journal sources but on a larger scale, encompassing 5,190 articles. As a result, this dataset was named Abstract-5190. From the same journal sources as the eye-tracking corpus, we selected academic papers with no English characters and abstracts containing more than 50 characters, excluding the original 320 eye-tracking samples, to construct the Abstract-5190 keyphrase extraction dataset, comprising 5190 papers. Detailed information about the dataset is provided in Table 6.

**Table 6. Summary of Keyphrase Extraction Corpora**

| Metrics | Dataset | |
|---|---|---|
| | Abstract-320 | Abstract-5190 |
| Number of abstracts | 320 | 5190 |
| Total number of sentences | 1215 | 15999 |
| Total number of characters | 64437 | 804210 |
| Number of characters (not repeated) | 1224 | 2099 |
| Coverage of eye-tracking features (character-level) | 94.69% | 54.65% |
| Number of key characters (not repeated) | 599 | 1250 |
| Key character eye movement feature coverage | 99.67% | 75.20% |

**Note:** The total number of characters includes Chinese characters and punctuation marks; key characters are those that appear in keyphrases.

## 3.5 Keyphrase Extraction Models

To better explore the effectiveness of character-level eye-tracking features in keyphrase extraction tasks, this study tested various models. The KPE models are categorized into two types: extraction models based on recurrent neural networks and those based on pre-trained language models.

### 3.5.1 Recurrent Neural Networks based KPE Models

Neural networks consist of multiple layers of neurons (or nodes) connected through weighted links. Each neuron receives input from the neurons in the previous layer, processes it using weights and activation functions, and produces output. We constructed four models: BiLSTM, BiLSTM+CRF, attention-based BiLSTM (Att-BiLSTM), and Att-BiLSTM+CRF. The simplest BiLSTM serves as the baseline model for our experiment. BiLSTM is a classic neural network model that captures contextual information from both directions, enhancing the understanding and modeling of dependencies in sequential data. Adding a CRF layer to the BiLSTM model allows for a more nuanced evaluation of label dependencies. Introducing an attention mechanism to BiLSTM and BiLSTM+CRF models enables dynamic adjustment of weight distribution by calculating the relevance of each hidden state to others, thus better capturing key information. Additionally, in this study, eye-tracking features were incorporated as external features into the BiLSTM and BiLSTM+CRF models. For the Att-BiLSTM and Att-BiLSTM-CRF models, eye-tracking features were used as attention indicators to guide sequence labeling.

### 3.5.2 Pre-trained Language Model-Based KPE Models

Pre-trained language models excel in language understanding and transfer learning, capturing character relationships more effectively than neural networks. We constructed keyphrase extraction models using BERT (Devlin et al., 2019), RoBERTa (Liu et al., 2019), and MacBERT (Cui et al., 2021). To further evaluate the effectiveness of eye-tracking features under more recent text-to-text pre-trained architectures, we incorporated two Chinese T5-style models. These two models were selected because the original Google T5 checkpoints showed tokenization and prediction artifacts on Chinese texts, such as underscore-like outputs. By contrast, Randeng-T5-784M is adapted to Chinese text processing and therefore provide more suitable T5-style baselines for Chinese academic KPE.

(1) *BERT*, based on BERT-base-chinese, includes 12 Transformer layers, 12 Attention Heads, and a hidden dimension of 768. It was pre-trained using Masked Language Model (MLM) and Next Sentence Prediction (NSP) tasks.

(2) *RoBERTa* was implemented using the Chinese-RoBERTa-wwm-ext checkpoint released by HIT and iFLYTEK. This model follows the RoBERTa pre-training strategy, incorporates whole-word masking (WWM) for Chinese text, and removes the next sentence prediction (NSP) objective used in BERT.

(3) *MacBERT,* based on chinese-macbert-base, modifies the standard masked language

modeling (MLM) objective into a correction-style task by replacing selected tokens with semantically similar words and predicting the original tokens. The model also adopts whole-word masking (WWM) and N-gram masking to better capture Chinese contextual representations.

(4) *Randeng-T5*[5]. Randeng-T5-784M is a larger Chinese text-to-text model based on the mT5-large architecture, with approximately 784M parameters, making it comparable in scale to T5-large.

# 4. Results Analysis

This section outlines the experimental parameters and evaluation metrics, analyzing the impact of different eye-tracking features and their combinations on keyphrase extraction (KPE) performance. In addition, a case analysis is provided visually demonstrates the enhancement in keyphrase extraction due to eye-tracking features.

## 4.1 Evaluation Method

The parameter settings can be found in Appendix A.3. The evaluation compares the $F_1$ scores of different models when extracting 3, 5, and 10 keyphrases. The calculation formula is as follows:

$$P = \frac{TP}{TP + FP} \tag{2}$$

$$R = \frac{TP}{TP + FN} \tag{3}$$

Where, *TP* represents the number of correctly predicted positive examples; *FP* indicates the number of false positives, where predictions are positive but actual values are negative; *FN* denotes false negatives, where predictions are negative but actual values are positive. The $F_1$ score is calculated as follows:

$$F_1 = \frac{2 \times P \times R}{P + R} \tag{4}$$

It is important to note that in academic texts, keyphrases present in the abstract do not cover all the keyphrases marked by the author. In this study, true positives only include keyphrases that appear in the abstract, excluding those not present in the abstract.

[5] https://huggingface.co/IDEA-CCNL/Randeng-T5-784M

## 4.2 Analysis of Experiments with single ET feature

Webcam-based eye-tracking offers a low-cost, flexible, and scalable alternative for KPE experiments. It allows customized, task-relevant datasets, supports data collection in diverse settings, better reflects real-world reading behavior, and enhances reproducibility and ecological validity compared to traditional eye-tracking or open-source cognitive corpora.

**Table 7. $F_1$ Scores of Models with Single Eye-Tracking Features (%)**

| KPE Model | ET Feature | Abstract-320 | | | Abstract-5190 | | |
|---|---|---|---|---|---|---|---|
| | | $F_1$@3 | $F_1$@5 | F1@10 | $F_1$@3 | $F_1$@5 | $F_1$@10 |
| BiLSTM | None | 18.99 | 18.90 | 19.32 | 23.81 | 22.78 | 21.38 |
| | FFD | 20.28 | 20.44 | 20.47 | **26.53** | **25.05** | 22.25 |
| | FN | **20.82** | **21.39** | **21.35** | 26.10 | 24.62 | **22.48** |
| | TFD | 20.08 | 21.32 | 20.66 | 24.58 | 23.87 | 22.09 |
| BiLSTM+CRF | None | 26.17 | 26.40 | 26.71 | 37.37 | 37.93 | 37.28 |
| | FFD | **28.20** | **28.38** | 28.34 | 38.21 | **38.49** | **38.04** |
| | FN | 27.77 | 28.18 | 28.20 | **38.28** | 38.00 | 37.63 |
| | TFD | 27.59 | 28.17 | **28.44** | 37.80 | 38.38 | 38.03 |
| Att+BiLSTM | None | 20.81 | 20.74 | 20.44 | 24.40 | 23.15 | 20.91 |
| | FFD | **22.57** | **22.64** | **22.26** | **26.52** | **25.07** | **22.77** |
| | FN | 22.30 | 22.51 | 21.83 | 24.42 | 23.21 | 21.48 |
| | TFD | 23.01 | 22.91 | 22.85 | 25.14 | 23.38 | 21.84 |
| Att+BiLSTM +CRF | None | 22.89 | 23.95 | 23.63 | 33.65 | 33.80 | 31.40 |
| | FFD | **25.91** | **25.81** | **25.45** | **35.58** | **36.05** | 33.57 |
| | FN | 25.22 | 25.41 | 24.67 | 35.57 | 35.73 | **34.03** |
| | TFD | 25.35 | 25.66 | 25.27 | 35.13 | 35.47 | 33.34 |
| BERT | None | 37.05 | 37.04 | 36.90 | 40.57 | 40.54 | 40.48 |
| | FFD | **39.39** | **39.87** | **39.70** | **42.63** | **41.21** | **41.14** |
| | FN | 38.98 | 39.76 | 39.67 | 40.66 | 40.88 | 40.91 |
| | TFD | 38.93 | 39.63 | 39.35 | 40.39 | 40.65 | 40.69 |
| MacBERT | None | 39.31 | 39.44 | 39.38 | 42.39 | 42.65 | 42.55 |
| | FFD | 40.36 | 40.40 | 40.23 | **43.58** | 42.85 | 42.74 |
| | FN | 40.48 | 40.50 | 40.34 | 43.06 | **43.09** | **43.11** |
| | TFD | **40.54** | **40.58** | **40.41** | 43.3 | 43.21 | 43.11 |
| RoBERTa | None | 39.77 | 39.70 | 39.63 | 42.65 | 42.76 | 42.73 |
| | FFD | 41.35 | 41.14 | 41.03 | 43.78 | 44.36 | 44.28 |
| | FN | 41.37 | 41.26 | 41.15 | **43.88** | **44.51** | **44.38** |
| | TFD | **41.65** | **41.89** | **41.79** | 43.06 | 43.05 | 42.97 |

**Note:** Bold indicates the ET feature achieving the optimal $F_1$ under a specific KPE model.

As shown in Table 7, we investigated the impact of individual eye-tracking features (FFD, FN, and TFD) on Chinese LIS abstract-based KPE by integrating them into different KPE models and comparing against baselines without eye-tracking information. In view of the small sample size of the Abstract-320 dataset, one split will make the performance test susceptible to chance fluctuations; hence, the adoption of five-fold cross-validation provides a more stable and reliable estimate of model performance. By contrast, the Abstract-5190 dataset, because of its larger scale, yields stable evaluation results from a single train-test split, rendering the adoption of five-fold cross-validation unnecessary while minimizing computational requirements.

From the experimental results presented in Table 7, it is evident that models achieved higher performance on Abstract-5190, with ET features showing that ET features improve performance across both datasets. ET signals exhibit strong consistency in both attention-based and BERT-based models, attaining optimal performance when combined with FFD features. In BiLSTM-based models, FFD and FN features perform most favorably. In MacBERT-based models, the contribution of ET features is comparatively modest; conversely, RoBERTa attains peak KPE performance on the large-scale Abstract-5190 dataset when integrated with FN, and achieves optimal results on the smaller Abstract-320 dataset when combined with TFD features.

On the smaller-scale dataset (Abstract-320), a comparison of different KPE models reveals that FFD achieves the most pronounced improvement on the BERT model, reaching an enhancement of 2.83%. In contrast, on the larger-scale dataset (Abstract-5190), as the volume of training samples increases, the MacBERT model—which underperformed on the smaller dataset—demonstrates a more substantial improvement. RoBERTa, which exhibited the best performance on the small dataset, also experiences further gains, attaining its optimal performance of 44.51% when combined with FN. Collectively, these findings indicate that eye-tracking features contribute positively to KPE model performance, although the magnitude of improvement depends on the model architecture and dataset size. Notably, when data is limited, early-processing features such as FFD offer greater informational gain, whereas with abundant data, cumulative-processing features like TFD better support deep semantic modeling.

### 4.3 Analysis of Experiments with combined ET features

In previous experiments, we confirmed the positive effects of individual ET features on KPE performance. To comprehensively assess the combined effects of multiple eye-tracking features on extraction tasks, this study evaluated three feature combinations, detailed in Table 8. To better visualize the impact of different feature combinations on $F_1$ scores, we plotted comparative charts of $\Delta F_1@5$ across models (Refer to Figure 4 and Figure 5).

**Table 8. $F_1$ Scores of Models Combining Multiple Eye-tracking Features (%)**

| Model | ET Features | Abstract-320 | | | Abstract-5190 | | |
|---|---|---|---|---|---|---|---|
| | | $F_1$@3 | $F_1$@5 | $F_1$@10 | $F_1$@3 | $F_1$@5 | $F_1$@10 |
| BiLSTM | None | 18.88 | 19.05 | 19.44 | 23.81 | 22.78 | 21.38 |
| | FFD+FN | 20.65 | 20.77 | 20.20 | 26.15 | 24.94 | 21.79 |
| | FN+TFD | **22.05** | **22.09** | **22.20** | 25.01 | 23.75 | 21.78 |
| | FFD+TFD | 21.27 | 21.17 | 21.34 | 24.17 | 22.76 | 21.29 |
| | FFD+FN+TFD | 21.32 | 21.42 | 21.31 | **27.47** | **25.73** | **22.50** |
| BiLSTM+ CRF | None | 26.17 | 26.40 | 26.71 | 37.37 | 37.93 | 37.28 |
| | FFD+FN | 27.94 | 28.37 | 27.93 | **38.49** | 38.79 | 38.03 |
| | FN+TFD | **28.74** | **29.02** | **29.01** | 37.90 | 38.31 | 37.89 |
| | FFD+TFD | 28.45 | 28.89 | **29.01** | 38.04 | **38.86** | **38.16** |
| | FFD+FN+TFD | 28.02 | 28.37 | 28.11 | 37.92 | 38.28 | 37.93 |
| Att-BiLSTM | None | 20.81 | 20.74 | 20.44 | 24.40 | 23.15 | 20.91 |
| | FFD+FN | 23.04 | 22.99 | 22.34 | 26.66 | 25.14 | 23.14 |
| | FN+TFD | **23.35** | **23.65** | **23.15** | **27.34** | **25.30** | **23.26** |
| | FFD+TFD | 22.01 | 22.72 | 21.62 | 22.98 | 22.18 | 21.32 |
| | FFD+FN+TFD | 22.67 | 22.66 | 22.35 | 25.80 | 24.23 | 22.41 |
| Att-BiLSTM+ CRF | None | 22.89 | 23.55 | 23.23 | 33.65 | 33.80 | 31.40 |
| | FFD+FN | 25.24 | 25.60 | 24.75 | 35.06 | 35.54 | 33.94 |
| | FN+TFD | **25.85** | **25.62** | **25.21** | **36.36** | **36.79** | **35.46** |
| | FFD+TFD | 24.60 | 25.32 | 24.95 | 34.81 | 35.29 | 32.70 |
| | FFD+FN+TFD | 25.33 | 25.42 | 25.06 | 35.87 | 35.73 | 33.62 |
| BERT | None | 37.05 | 37.04 | 36.90 | 40.57 | 40.54 | 40.48 |
| | FFD+FN | 38.65 | 38.89 | 38.64 | 42.03 | 41.78 | 41.71 |
| | FN+TFD | 39.27 | 39.29 | 39.04 | **42.06** | **41.93** | **41.83** |
| | FFD+TFD | **39.45** | **39.41** | **39.31** | 41.05 | 41.85 | 41.81 |
| | FFD+FN+TFD | 38.57 | 38.69 | 38.73 | 41.67 | 41.71 | 41.67 |
| MacBERT | None | 39.31 | 39.44 | 39.38 | 42.39 | 42.65 | 42.55 |
| | FFD+FN | 40.00 | 40.37 | 40.20 | 43.1 | 43.15 | 43.1 |
| | FN+TFD | 40.42 | 40.09 | 40.13 | 43.01 | 42.88 | 42.85 |
| | FFD+TFD | **40.52** | **40.81** | **40.64** | **44.16** | **43.71** | **43.65** |
| | FFD+FN+TFD | 40.36 | 40.03 | 39.85 | 43.56 | 43.65 | 43.57 |
| RoBERTa | None | 39.77 | 39.70 | 39.63 | 42.65 | 42.76 | 42.73 |
| | FFD+FN | 41.65 | 41.82 | 41.71 | **44.72** | **44.65** | **44.67** |
| | FN+TFD | 41.79 | 41.77 | 41.73 | 42.23 | 42.51 | 42.44 |
| | FFD+TFD | **42.26** | **42.22** | **42.17** | 43.43 | 42.58 | 42.48 |
| | FFD+FN+TFD | 41.46 | 41.78 | 41.68 | 43.53 | 43.87 | 43.89 |

The findings indicate that joint feature effects are heterogeneous and dependent on the model structure and dataset scale. As shown in Table 8, on Abstract-320, the best performing deep learning model is BiLSTM+CRF, while the BiLSTM model gets the greatest improvement (3.04%) when incorporating FN+TFD. Better extraction results are achieved by BERT pre-trained language models, with the top performing results being from the RoBERTa-based KPE model. Furthermore, adopting ET signals results in significant improvement in terms of performance for RoBERTa as well as BERT models. Conversely, on the larger Abstract-5190 dataset, combining FN+TFD and FFD+FN+TFD results in more stable and consistent improvement, particularly noticeable in the BiLSTM and in the Att-BiLSTM+CRF models, in which such improvement is almost 3%.

In the Abstract-5190 dataset, the combination of FFD+FN demonstrated a more reliable enhancement effect in BiLSTM-based KPE models. Although the FN+TFD pairing reached its peak improvement under the ATT_BiLSTM+CRF configuration, its enhancement magnitude, like that of FFD+FN, remained unstable. Both combinations occasionally exhibited negative impacts on KPE performance (particularly in the ATT_BiLSTM and RoBERTa models). For BERT and its variants, eye-tracking signals showed the most consistent effects within the BERT model, where FFD+FN again proved to deliver steady performance gains. Overall, the FFD+FN and FFD+FN+TFD combinations contributed more stable and dependable improvements to academic KPE tasks.

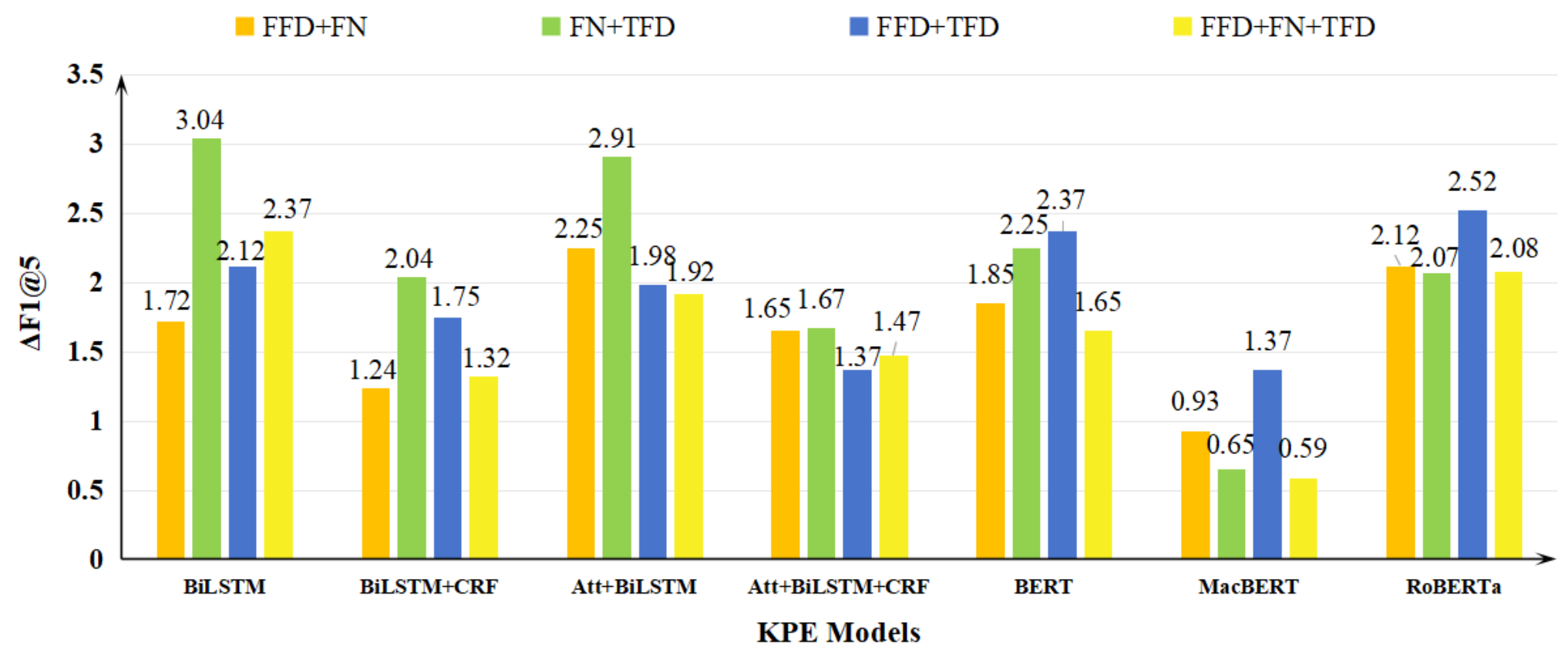

**Figure 4. Comparison of Δ $F_1$@5 of combined eye movement feature (Abstract-320)**

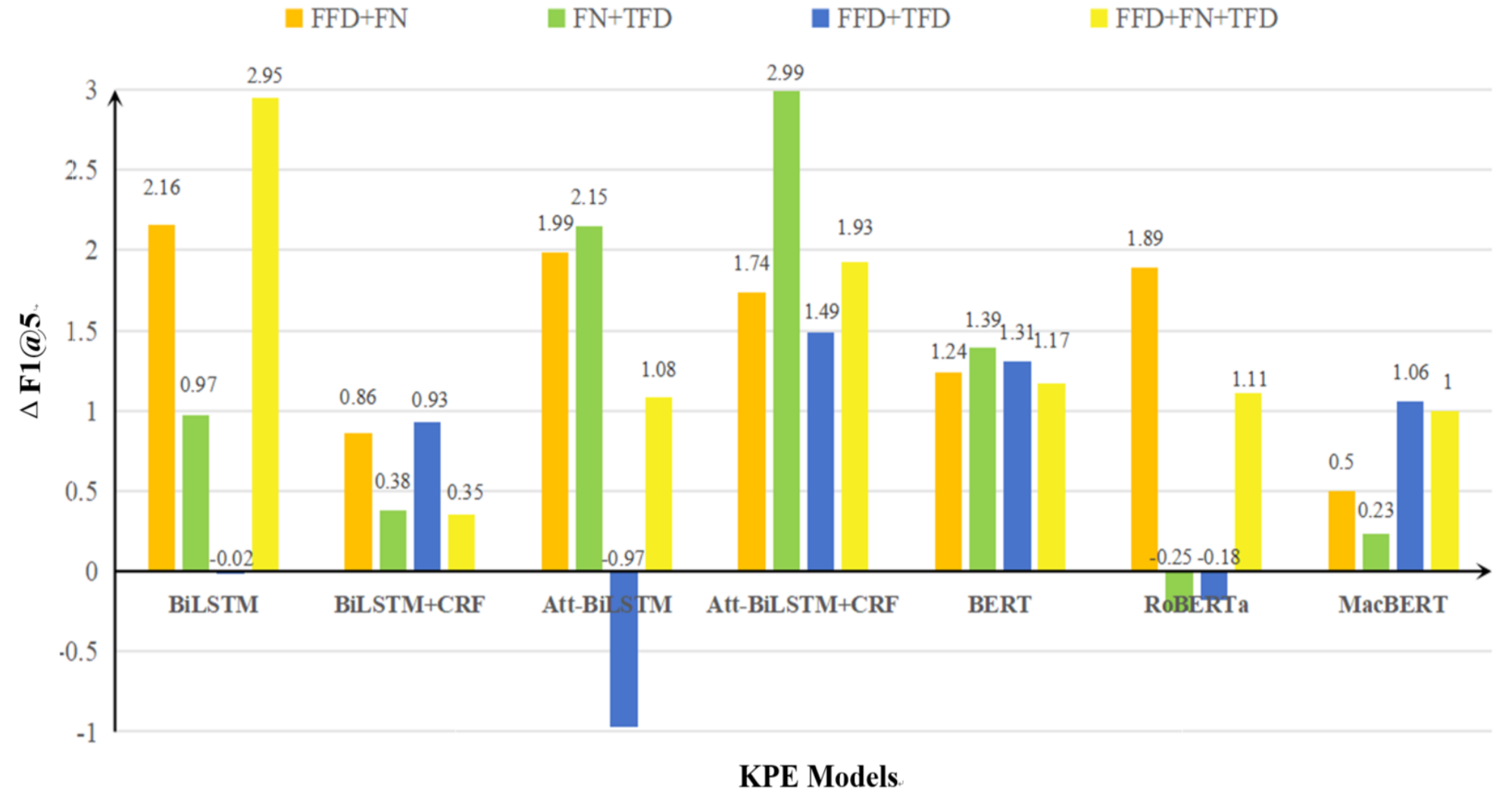

**Figure 5. Comparison of Δ $F_1$@5 of combined eye movement feature(Abstract-5190)**

As a whole, these findings suggest the emergence of some common trends: (1) the influence of ET feature combinations largely depends on the structure of the model, with models improved by CRF and pre-trained models revealing a higher potential to take combined signals in their entirety; (2) FN+TFD and FFD+TFD achieve more pronounced enhancements for KPE on smaller datasets, while FFD+FN and FFD+FN+TFD demonstrate more stable improvement effects on larger-scale datasets; and (3) although the three-feature combination of FFD+FN+TFD does not yield further optimization, it continues to maintain its reinforcing effect on document-level performance.

## 4.4 Significance Testing of ET features on Abstract-320 KPE Dataset

As a means of compensating for incidental benefit from ET features on the limited Abstract-320 dataset, we utilized a five-fold cross-validation process. On each pass through, we used a two-tailed paired t-test to assess statistical significance, with p-values as the primary measure of assessment. The p-values are reported in Table 9, and the corresponding visualization is shown in Figure 6.

The results show that ET features exert statistically significant effects across multiple KPE architectures, although the degree of significance varies by model type and feature combination. In BiLSTM-based models, FFD and FN show significant effects in several settings, suggesting that both early fixation duration and fixation frequency can provide useful cognitive signals for sequence labeling. Among combined features, FN+TFD and FFD+TFD show more consistent effects, indicating that fixation frequency and fixation duration capture complementary aspects of reading behavior.

For BERT-family models, the significance patterns suggest that ET features can complement contextual semantic representations, although the degree of contribution differs across models. BERT and MacBERT show stronger significance for several individual and combined features, especially FFD, FN, and TFD-related settings. In contrast, RoBERTa shows weaker significance overall, with only a few feature settings reaching statistical significance. This indicates that the usefulness of ET features is not uniform across pre-trained language models and may depend on how each model encodes contextual information.

The results of Randeng-T5 further support the robustness of ET features under a stronger Chinese text-to-text pre-trained architecture. For Randeng-T5, duration-based features such as FFD and TFD show highly significant effects, while FN alone is less stable. FN+TFD and FFD+TFD exhibit the strongest significance, whereas the three-feature combination does not produce the strongest effect.

This suggests that paired combinations of complementary cognitive signals may be more effective than simply adding all available ET features.

Overall, the significance analysis confirms that ET features provide useful supplementary signals for KPE on the Abstract-320 dataset. Among individual features, fixation-duration indicators show relatively stable effects, while among combined features, FN+TFD demonstrates the most consistent significance across model types. These findings support the value of human reading behavior for cognitively grounded KPE and further suggest that future work should explore more refined feature-fusion strategies for strong pre-trained and text-to-text architectures.

**Table 9. P-values of Multiple Eye-tracking Features in various KPE models**

| ET Features / KPEModels | FFD | FN | TFD | FFD+FN | FN+TFD | FFD+TFD | FFD+FN+TFD |
|---|---|---|---|---|---|---|---|
| **BiLSTM** | 0.0465* | 0.0151* | 0.1438 | 0.0228* | 0.0057** | 0.0036** | 0.0152* |
| **BiLSTM +CRF** | 0.0603 | 0.0273* | 0.0958 | 0.0393* | 0.0076** | 0.0970 | 0.0348* |
| **ATT_BiLSTM** | 0.1260 | 0.0145* | 0.0060** | 0.0261* | 0.0023** | 0.0037** | 0.0097** |
| **ATT_BiLSTM +CRF** | 0.0462* | 0.0039** | 0.0676 | 0.0365* | 0.0032** | 0.0197* | 0.0125* |
| **BERT** | 0.0002** | 0.0011** | 0.0201* | 0.0090** | 0.0058** | 0.0168* | 0.1151 |
| **MacBERT** | 0.0407* | 0.0037** | 0.0022** | 0.1510 | 0.0325* | 0.0548 | 0.0192* |
| **RoBerta** | 0.1796 | 0.1411 | 0.0493* | 0.0594 | 0.0722 | 0.0248* | 0.1291 |
| **Randeng-T5** | 0.0023** | 0.0785 | 0.0099** | 0.0408* | 0.0006** | 0.0042** | 0.0122* |

**Note:** * $p < 0.05$; ** $p < 0.01$.

**Figure 6.** P value for individual and combined ET features in different KPE models

## 5. Conclusion and Future Works

Through this webcam-based eye-tracking setup, we successfully captured eye-movement data during the reading of Chinese academic abstracts, thereby constructing the Chinese Library and Information Science Eye-Tracking Corpus (CLIS-ET). By integrating eye-tracking features into keyphrase extraction (KPE) models, we achieved notable performance gains, with the combination of fixation number and total fixation duration (FN + TFD) yielding the most significant improvement in the Att-BiLSTM + CRF model.

Compared with traditional eye-tracking methods that rely on expensive equipment and controlled laboratory environments, our approach uses ordinary webcams and open-source frameworks, substantially reducing experimental and deployment costs while maintaining reliable fixation accuracy. This demonstrates advantages in hardware accessibility and deployment flexibility across diverse research contexts.

The proposed framework not only enhances KPE performance from a cognitive perspective but also may inform future user-centered applications such as intelligent academic reading systems, personalized literature recommendation, and educational assistance platforms. By incorporating real reader attention patterns into text comprehension and information retrieval, it contributes to building NLP systems that better align with human cognition, thereby improving the interpretability and human-centeredness of human–computer interaction.

Moreover, the developed data collection framework supports large-scale, synchronized eye-tracking acquisition and can serve as a foundation for cognitive NLP research in future extensions to other academic domains and participant groups. In future work, we plan to expand the academic eye-tracking corpus, improve tracking quality and validation procedures, and further explore how integrating psychological insights into computational models can advance understanding of cognitive and learning processes in reading-based NLP tasks.

# Ethics Statement

This study utilized publicly available data from three Chinese academic journals to design an eye-tracking reading experiment. Before the experiment, all participants were given an informed consent form (Appendix A.3) to read and sign. The personal information of participants was safeguarded and is not included in the publicly accessible data. The eye-tracking experiment was conducted following principles of accuracy and transparency, with no subjective influence in the data recording or processing.

## Acknowledgment

This work is supported by the National Natural Science Foundation of China (Grant No. 72074113) and the Major Projects of National Social Science Fund (Grant No. 25&ZD298).

## Author conflict statement

The author(s) declared no potential conflicts of interest with respect to the research, authorship, and/or publication of this article.

# Appendix

### A.1 Comparison of Current Eye-tracking Collection Tools

According to Table A1, this comparison is intended to highlight differences in hardware requirements and deployment accessibility. Approximate costs are based on publicly available vendor information, quotations reported in the literature, and market estimates. Actual prices vary depending on hardware configuration, software licenses, accessories, institutional agreements, and purchase date. The table compares hardware characteristics and deployment requirements only and does not evaluate tracking accuracy, precision, or data quality.

**Table A1** Comparison of Representative Eye-Tracking Systems

| Device | Hardware | Sampling Rate | Environment | Accessibility | Approximate Cost (USD) |
|---|---|---|---|---|---|
| EyeLink 1000 Plus | Dedicated infrared tracker | up to 2000 Hz | Laboratory | Specialized hardware required | $25,000–40,000+ |
| EyeLink Portable Duo | Portable infrared tracker | up to 2000 Hz | Laboratory / Field | Specialized hardware required | $30,000–50,000+ |
| Tobii Pro Spectrum | Screen-mounted infrared tracker | up to 1200 Hz | Laboratory | Specialized hardware required | $20,000–40,000+ |
| Tobii Pro Fusion | Screen-mounted infrared tracker | up to 250 Hz | Laboratory/ Office | Specialized hardware required | $5,000–15,000+ |
| SearchGazer | Webcam-based tracker | 30–60 Hz | Laboratory/ Classroom/ Remote | Consumer-grade hardware | $0–200 |

### A.2 Experimental Environment and Participants

The SearchGazer library is an eye-tracking tool that employs the getUserMedia/Stream API to access webcams for webcam-based eye-tracking data collection. Written in JavaScript, SearchGazer can be integrated into web-based experimental interfaces for remote eye-tracking research with just a few lines of code. It has a maximum sampling rate of 60Hz, with an interval of approximately 16.67ms between samples, sufficient to identify fixation segments ranging from 50ms to 1500ms (Rayner, 1998). In the eye movement data collection platform, reading materials are displayed in short sentences at the center of the browser screen, ensuring full display on common laptop screens. For consistency, the font size is set to 50px, with 10px letter spacing, a maximum of 19 characters per line, and a maximum of two lines with a line spacing of 100px (Refer to Fig. 2).

The eye-tracking experiment followed a rigorous protocol, beginning with participants signing informed consent forms to confirm their understanding and agreement. Participants then underwent a training session, which included a video demonstration and explanation of the annotation guidelines to ensure they understood the experimental procedures. Following training, participants configured the experimental environment, setting up their computers, webcams, and software according to the Eye-Tracking Experiment Arrangement Document. A pre-experiment phase, utilizing ten reading materials, allowed participants to familiarize themselves with the process before the formal data collection began. The formal experiment consisted of four key stages: initial webcam calibration, secondary mouse-based gaze point calibration, natural reading (allowing participants to freely switch between texts), and a post-reading comprehension assessment. Reading comprehension was assessed using a post-reading test, with accuracy measured as the percentage of correct answers.

Ten graduate students or senior undergraduates from the Library and Information Science field, all native Chinese speakers (detailed in Table A2), participated in the study. Effective eye movement points were defined as the number of recorded eye movement coordinates falling within the reading material's coordinate range. This experiment was conducted in full compliance with ethical guidelines, and no ethical concerns were identified.

**Table A2** Summary of 10 subjects

| No. of participants | ID | Age | Sex | Major | Total durations/s | Average durations/s | Effective eye-tracking point | Accuracy |
|---|---|---|---|---|---|---|---|---|
| 1 | 202301 | 23 | Female | LIS | 21,515.06 | 67.23 | 366037 | 95.63% |
| 2 | 202302 | 22 | Female | LIS | 17,596.45 | 54.99 | 185110 | 90.00% |
| 3 | 202303 | 23 | Male | LIS | 14,774.43 | 46.17 | 137372 | 93.12% |
| 4 | 202305 | 24 | Male | LIS | 17,814.57 | 55.67 | 446088 | 92.81% |
| 5 | 202306 | 23 | Male | LIS | 15,180.98 | 47.44 | 206053 | 90.94% |
| 6 | 202308 | 24 | Male | LIS | 15,030.68 | 46.97 | 160350 | 92.19% |
| 7 | 202309 | 24 | Male | LIS | 15,726.64 | 49.15 | 186662 | 97.19% |
| 8 | 202304 | 22 | Male | IMIS | 12,985.41 | 40.58 | 314235 | 92.50% |
| 9 | 202307 | 21 | Male | IMIS | 13,643.29 | 42.64 | 210735 | 92.50% |
| 10 | 202310 | 21 | Female | IMIS | 19,849.05 | 62.03 | 375779 | 96.88% |

**A.3 Parameter Settings**

This study utilizes the Abstract-320 and Abstract-5190 datasets as test sets. Given that the maximum length of the summary text is approximately 500 characters, we set the max_length to 512. For the smaller Abstract-320 dataset, five-fold cross-validation is employed to reduce random bias. The Abstract-5190 dataset is divided into training and test sets in a 4:1 ratio. Due to the different sizes of the two datasets, the learning rates and training epochs required for model fitting also differ. For the smaller Abstract-320 dataset, the training epochs and learning rates are set to 30 and 0.01, respectively, for the BiLSTM and BiLSTM+CRF models; for models based on attention mechanisms, the training epochs are 65 and the learning rate is 0.005. For the Abstract-5190 dataset, the training epochs for the aforementioned four models are 30, with a learning rate of 0.003. In the three pre-trained language models, the training epochs for the Abstract-320 and Abstract-5190 datasets are 10 and 8, respectively, with a learning rate of 5e-5.

The main experiments were conducted on a high-performance computing server equipped with NVIDIA A100 GPUs. During the revision stage, Randeng-T5-784M was added as an additional Chinese text-to-text model to further examine whether eye-tracking features remain useful under a stronger pre-trained architecture. Because access to the original A100 computing environment was no longer available during the supplementary experiments, Randeng-T5-784M was trained on a workstation equipped with an NVIDIA RTX 4090 GPU. The Randeng-T5 experiments were conducted on the Abstract-320 dataset using five-fold cross-validation. The maximum sequence length was set to 512, the learning rate was set to 5e-5, and AdamW was used as the optimizer. The model was trained for up to 20 epochs with early stopping, where the patience value was set to 5.